# A Novel Frame Identification and Synchronization Technique for Smartphone Visible Light Communication Systems Based on Convolutional Neural Networks

Vaigai Nayaki Yokar[1], Hoa Le-Minh[1], Xicong Li[1], Wai Lok Woo[2], Luis Nero Alves[3], Stanislav Zvanovec[4] , Tran The Son[5] and Zabih Ghassemlooy[1]
[1] Optical Communications Research Group, Department of Mathematics, Physics and Electrical Engineering, Northumbria University, Newcastle Upon Tyne, United Kingdom.
[2]Department of Computer and Information Sciences, Northumbria University, Newcastle Upon Tyne, United Kingdom.
[3]Instituto de Telecomunicações and the Departamento de Electrónica, Telecomunicações e Informática, Universidade de Aveiro, 3810-193 Aveiro, Portugal.
[4]Department of electromagnetic field, Czech Technical University, Praha, Czech Republic.
[5]Faculty of Computer Engineering and Electronics, Vietnam-Korea University of Information and Communication Technology-University of Da Nang, Da Nang, Vietnam.

Corresponding author: Vaigai Nayaki Yokar (e-mail: vaigai.yokar@bristol.ac.uk).

"This work was supported by the European network on future generation optical wireless communication technologies under the Cost Action – CA19111 (New focus) and Instituto de Telecomunicações and the Departamento de Electrónica, Telecomunicações e Informática, Universidade de Aveiro."

**ABSTRACT** This paper proposes a novel, robust, and lightweight supervised Convolutional Neural Network (CNN)-based technique for frame identification and synchronization, designed to enhance short-link communication performance in a screen-to-camera (S2C) based visible light communication (VLC) system. Developed using Python and the TensorFlow Keras framework, the proposed CNN model was trained through three real-time experimental investigations conducted in Jupyter Notebook. These experiments incorporated a dataset created from scratch to address various real-time challenges in S2C communication, including blurring, cropping, and rotated images in mobility scenarios. Overhead frames were introduced for synchronization, which leads to enhanced system performance. The experimental results demonstrate that the proposed model achieves an overall accuracy of approximately 98.74%, highlighting its effectiveness in identifying and synchronizing frames in S2C VLC systems.

## I. INTRODUCTION

Visible light communications (VLC) is a subset of optical wireless communication technology that typically transmits information by modulating the intensity of light sources, such as lasers, light emitting diodes (LEDs), organic LEDs (OLEDs), screens and panels, in the visible spectrum (380 – 780 nm) [1] [2]. The information reception can be realized by a photodiode or an image sensor . The technology of utilizing an image sensor for visible light communication is also known as optical camera communication (OCC) because the image sensor is the core component of a camera, and the image sensor and camera are often used interchangeably in the literature[3] [4]. Given the widespread presence of built-in cameras on billions of [5]smartphones worldwide, OCC offers an innovative and cost-effective software solution for short-range communication that eliminates the need for additional hardware [6] [7]. OCC enables smartphones to communicate seamlessly with one another even when the wireless connections, e.g., WiFi and Bluetooth, are all disabled or banned in some scenarios, since the display can act as a light source (transmitter, Tx) for transmitting data and the camera can be used for data reception (receiver, Rx). This is known as screen-to-camera communications (S2C) . Motivated by OCC's potential and market opportunities, the IEEE 802.15.7 standard, released in 2015, is the first to recognize OCC as an industry standard [8].

The following research works in S2C systems have been carried out. In [9], a cellular overlap-based resource allocation technique (COBRA) was introduced for S2C to enhance data transmission in bandwidth-limited RF communication systems. This method utilized a specialized two-dimensional barcode to optimize resource allocation in 2010. The COBRA S2C has been demonstrated with a throughput of up to 225 kbps, laying the groundwork for improving data transmission in smartphone communication. In practice, COBRA requires a high-quality Rx as it is restricted by frame rate and light sensitivity. Followed by, visible light based near-field communication for smartphones (VINCE) [10] introduced a grayscale-based barcode transmission, in which Tx was divided into four areas and can be detected by Arduino signal processor from the receiving end. In [11], another near field communication based on visible light (NECAS) achieved better performance by introducing a colormap scheme. However, both VINCE and NECAS are sensitive to ambient noise. By contrast, robust application driven visual communication using color barcodes (RainBar) was proposed in [12] and achieved a higher data throughput of up to 958 kbps. Nevertheless, RainBar is vulnerable to lighting conditions and may not be suitable for transmitting long consecutive data. The authors in [13] introduced an automatic controlling system called SoftLight that utilizes an algorithm to adjust the Tx screen's brightness and contrast and effectively improves the visibility of a barcode or other visual markers at the Rx partner smartphone. The authors demonstrated robust and stable performance in various environments. Furthermore, researchers in [14] proposed a system Tetra-Transmission (TETRIS), which achieved a throughput of 311.22 kbps with a transmission accuracy of 90%. However, TETRIS is susceptible to ambient light noise and demands significant processing power. In speeded up robust features based S2C communication system [15], the authors demonstrated a system with projective transformation to eliminate perspective distortions caused by device displacement. Nevertheless, blurring was still one of the major challenges in such communication systems. It is worth noting that, frame identification and synchronization are the most common challenges in the above-mentioned studies and have not yet been thoroughly addressed. The asynchronous nature of data transmission between the Tx and Rx in these S2C systems, combined with the lack of synchronisation mechanisms between them, can result in significant error rates if the problem is not addressed properly.

Some research works focusing on frame synchronization have been carried out in VLC systems. In [16], asynchronous LED array detection was outperformed by the system using the synchronous LEDs. In this scheme, based on the sync-LEDs, the receiver identifies the starting point of an on-off keying (OOK)-modulated data sequence and calibrates the distorted image snapshots. In [17], the researchers proposed a LightSync approach, in which transmitted and lost frames can be monitored and recovered by inter-frame ensure coding technique. In [18], the research showed that transmitting a preset preamble allows synchronization with the start bit of the VLC data packet. In their proposed scheme, the authors also showed that the periodic nature of packet transmissions allows the Rx to synchronize with the Tx. In [19], HiLight concept, a new form of real-time screen-camera communication without showing any coded images (e.g., barcodes) for off-the-shelf smart devices, was introduced. Nevertheless, the alternative approach of using machine learning (ML) models to overcome the frame identification and synchronization challenges with reduced computation power and time is not explored in the above studies. ML has gained popularity in recent years due to its remarkable capabilities in image processing and object identification [20]. In practice, a major challenge of using Convolutional Neural Network (CNN) in an S2C system is the constrain in processing power, battery, and memory resources of the smartphone as standard CNNs are computationally intensive and may require specialized hardware or software optimizations to run efficiently on these portable devices.

To address these challenges, we provide a novel CNN based S2C system. The proposed S2C system focuses on (*i*) enhancing the detection and classification accuracy of the transmitted codes, (*ii*) improving the system synchronization by robustly detecting and separating the overhead frame from data, and (*iii*) improving the system adaptability to different environmental conditions by introducing augmented data.

The rest of the paper is organized as follows. Section II describes the system design and is followed by a description of the proposed CNN model in Section III. The experimental setup, analysis and results are presented in Section IV. Section V concludes our work.

## II. SYSTEM DESIGN
### A. SYSTEM OVERVIEW
The concept of an S2C system is illustrated in Fig. 1(a), where the Tx phone displays a picture containing the information to be sent to its partner Rx phone. In Fig. 1(b), a detailed block diagram for both Tx and Rx is provided. The data stream $d_s(t) \in \{0,1\}$, at the Tx, is divided into multiple data frames $d_f(t)$, where each frame has multiple $M$-pixel×$N$-pixel cells [2] [15].

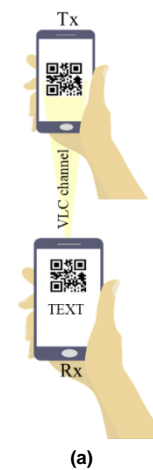

(a)

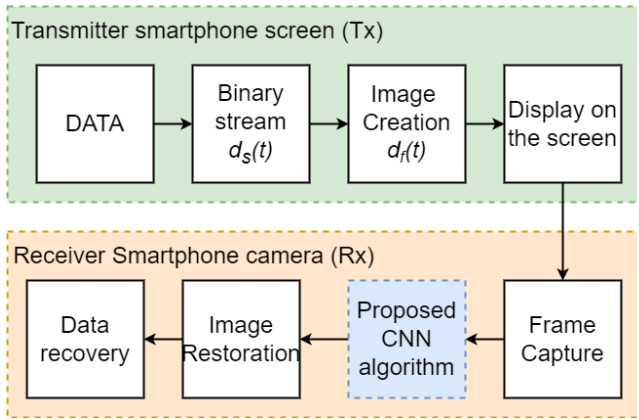

**Figure. 1.** Smartphone-based VLC system: (a) Tx and Rx pair, and (b) functional system block diagram

In OOK or color shift keying (CSK)-based systems, these cells are colored in white and black for '0's and '1's, respectively. A bit '1' is simply represented as an optical pulse that occupies the entire part of the bit duration while a bit '0' is represented by the absence of an optical pulse. For OOK modulation in S2C systems, non-return-to-zero (NRZ) line coding is often preferred due to its high spectrum efficiency. These frames are displayed in a predefined order on the Tx's screen. At Rx, the camera captures the successive frames and extracts the transmitted data encoded in these frames from identified regions of interest (ROI) using the proposed image restoration and CNN algorithms.

### B. FRAME IDENTIFICATION

In the proposed S2C system, each frame consists of a marking area and payload. The payload in a data frame contains the actual data, while an overhead frame contains a specific code for synchronization. At the Rx, each frame is converted back into a binary stream to recover the original data.

The proposed experimental investigation aims to enhance the performance of frame identification in the S2C systems by conducting three types of experiments using a CNN model. These experiments focus on (a) payload classification, (b) data frame detection, and (c) overhead frame detection. Conducting these three experiments have demonstrated significant improvement to S2C systems. Training the image recognition system with diverse data enhances its ability to distinguish between different types of transmission codes. This approach ensures accurate data transmission, robust signal detection, and effective synchronization, making the S2C system more resilient to various interferences and maintaining high accuracy.

### C. SYNCHRONIZATION

Synchronization between the Tx and Rx plays a crucial role in data recovery, thereby reducing errors and improving overall communication reliability. In practice, synchronization in the S2C system is always the most challenging task, since the Tx and Rx are independent in most systems and Rx needs to recover the timing information from the received data. Furthermore, Tx LEDs usually transmit data at a high frame per second (fps), while the camera receiving rate is typically much lower. For example, a Google Pixel 6 Pro can run at a frame rate of 60 fps and a screen refresh rate of 120 fps as shown in Fig. 2.

Also, varying distance between Tx and Rx can cause fluctuations in received strength alongside user mobility and signal interference can result in interference noise and fading. These factors contribute to a challenge in accurately detecting the overhead frames and achieving synchronization.

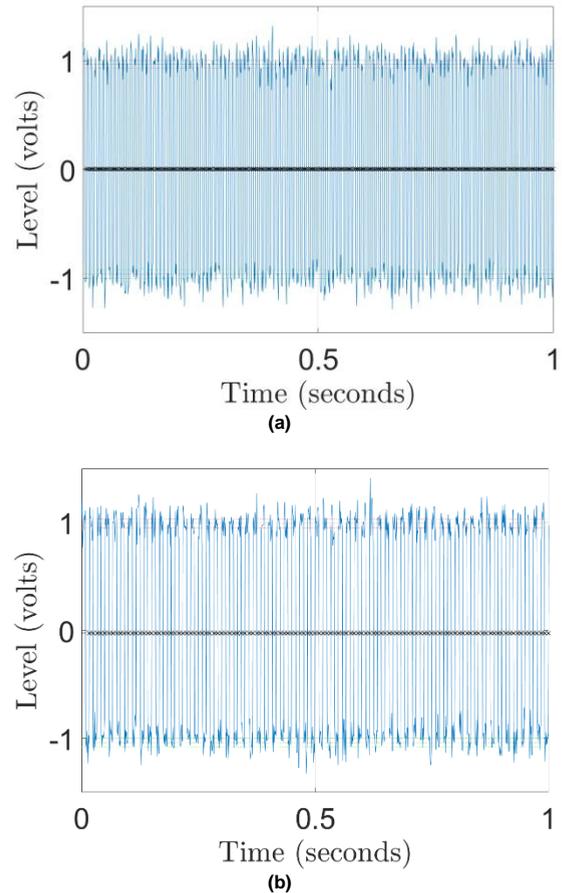

**Figure. 2.** Frame rate: (a) Tx (@120 fps), and (b) Rx (@60 fps).

The asynchronous transmission in S2C communication leads to (i) latency considerations, (ii) frame skipping or duplications, and (iii) resilience to timing variations. To address this issue, in this work, overhead frames will be transmitted periodically, which helps the Rx to identify the beginning of a stream of packets as well as to correct any time offset. This helps to ensure that the frames are correctly received and processed by the system, leading to more reliable and accurate communication between smartphones as represented in Fig.3.

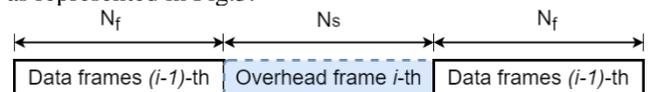

**Figure. 3.** S2C frame model

The traditional method to separate the random data frames from the overhead frames can be computationally extensive. To reduce computational power and system complexity, we propose a CNN model for frame identification and synchronization in S2C communication system.

## III. PROPOSED CNN ALGORITHM

CNN uses the convolutional process for extracting features from the input frames (or images). Followed by a set of filter/kernels are applied to the input images and feature maps are produced that highlight the different aspects of inputs, such as edges, patterns, and texture. The CNN approach is suitable in S2C as (i) it is very accurate in image recognition and classification, (ii) it minimizes computation time when compared to other NNs, (iii) CNN is specifically designed to leverage the spatial relationships and local patterns within images, and (iv) the network learns hierarchical representations of the input data, capturing both low-level features (e.g., edges) and high-level features (e.g., complex shapes).

This proposed supervised learning network is built using the following layers: i) 2D convolutional layer, ii) Max pooling layer, iii) ReLU (rectified linear unit), iv) Flatten and v) Fully connected layer as depicted in Fig. 4. The proposed CNN architecture is written in Python using the TensorFlow Keras framework. The proposed successive layers after the input are outlined as:

Layer 1: Convolutional layer with 32 filters, 3x3 kernel, ReLU, No padding.
Layer 2: One convolutional layer with 16 filters, 3x3 kernel, ReLU, No padding.
Layer 3: One max pooling layer with 2x2 pooling.
Layer 4: Flatten layer.
Layer 5: Fully connected layer with 128 neurons.
Layer 6: Output layer with 1 neuron (Logistic).

This structure is designed to achieve a unique balance between lightweight design and efficient performance, making it ideal solution for detection and classification in S2C system. The configuration prioritizes flexibility and speed without reducing efficiency and performance, setting it apart from conventional designs.

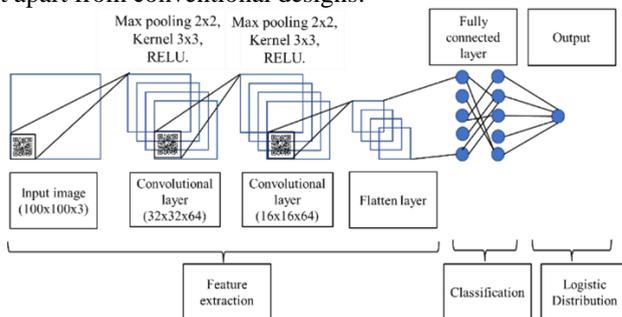

**Figure. 4.** Proposed CNN model.

Here, the CNN algorithm was created to classify $100 \times 100$ image size QR and ASCII codes. The image is convolved using three separate kernels, each of size $3 \times 3$, without striding in the convolution layer. A kernel size of "$3 \times 3$" is a common choice in CNNs, as it has been shown to be effective for a wide range of image processing tasks.

Following the convolution, the output "$24 \times 24$" pixel feature maps were created and subjected to ReLU activation function processing. By employing "$2 \times 2$" the maximum pooling technique, the size of the feature maps in the following layer has been reduced by half. Additionally, a "$2 \times 2$" pooling operation can help to introduce some degree of translation invariance into the model, as the pooling operation will produce the same output regardless of the exact location of the salient feature within the block. The receptive field in the maximum pooling layer is $2 \times 2$ with stride of $12 \times \mathbf{12}$ has further reduced feature map size to half. These feature maps were then flattened and passed through a fully connected layer, having neuron counts of 128. An optimizer is required to improve results of a CNN network where several layers are added one at a time. Optimizer aids in the creation of the output layer by the networks. Training and test results are required to analyze a task with a ML algorithm and to monitor that model. Although CNN offers a variety of optimizer techniques, we choose the "Adaptive moment estimation Optimizer" to minimize the loss, adaptive learning, efficient memory usage, and robustness to sparse gradients. After building the entire network, the entire model was compiled using the Adam optimizer to produce results. To train the model and evaluate the outcomes, we used 20 epochs.

## IV. EXPERIMENTAL SETUP, RESULTS AND ANALYSIS

The S2C setup includes two Google Pixel 6 Pro smartphones which are used as the Tx and the Rx. The key parameters are given in Table I.

TABLE I
KEY PARAMETERS OF THE SYSTEM

| Parameters | Description | Values |
|---|---|---|
| **Tx** | Smartphone | GooglePixel 6 Pro |
| | Display | OLED type |
| | Display size | 6.41'' |
| | Refresh rate | 120 Hz |
| *Frame size* $M \times N$ | Size of the frame | $100\ pixel \times 100\ pixel$ |
| *n* | Number of bits | 32000 |
| *char* | Number of chars | 4000 |
| **Rx** | Smartphone | GooglePixel 6 Pro |
| | Number of camera pixels | 50 MP |
| | Aperture | f/1.85 |
| | Sensor Model | CMOS |
| | Chip | Second-generation Tensor chip |
| | Shutter Speed | 1/8000 |
| | Camera frame rate | 60 fps |
| | Camera focus | Auto focus |
| **Channel** | **Visible light channel** | **380-780 nm** |
| *d* | Distance between the Tx and Rx | 20 cm |
| *t* | Tilt angle | $0^0$ |
| *r* | Rotation angle | $0^0$ |
| | Communication link | Line-of-sight |

The proposed supervised model is trained using a diverse range of datasets, as summarized in Table II. This unique dataset, built entirely from scratch, consists of four types of QR and ASCII-labeled images, with 1,000 images for each type. Several preprocessing steps were applied to the datasets before they were introduced into the network. Crop and rotate based augmented images generated from the datasets are obtained for the training. We utilized 75% of the dataset for CNN training and validation, reserving the remaining 25% for testing.

TABLE II
KEY LABELLED IMAGES OF DATASET

| Label | Type | Payload | Crop | Rotate |
|---|---|---|---|---|
| Data frame 1 ($d_{f1}$) | QR | | | |
| Data frame 2 ($d_{f2}$) | QR | | | |
| ASCII frame ($a_f$) | ASCII | | | |
| Overhead frame ($o_f$) | QR | | | |

### A. FRAME IDENTIFICATION

The detection and classification accuracy of the proposed supervised model is essential in the S2C system due to (i) Reliable communication, (ii) error prevention, (iii) performance evaluation, (iv) user experience and (v) system optimization. The following investigation to check the detection and identification accuracy of the model was carried out using three different experimental investigations.

1) Payload detection

The first experiment aims to classify different payloads within frames using a CNN model trained to distinguish between two types of encoded text data ($d_{f1}$ and $d_{f2}$ in Table II), along with augmented data generated by rotation and cropping. This enables precise data reading and robust signal detection under varying illumination and interference conditions. The training process, as shown in Fig. 5, begins with an initial accuracy of less than 0.48%, eventually reaching 0.98 within 8 epochs. Similarly, the validation curve starts at 0.45% accuracy, achieving 0.98 after 7 epochs.

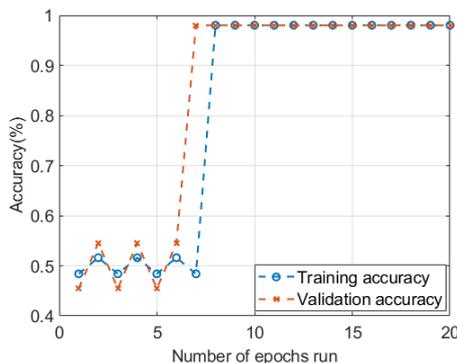

Figure. 5. Training and testing accuracy of payload detection.

2) Data frame detection

The second experiment focuses on ensuring the system exclusively recognizes data frames (QR codes), filtering out other barcodes like ASCII codes. The CNN model is trained on various barcodes to differentiate data frame, maintaining communication integrity by rejecting incorrect signals. This robust detection is effective even in high mobility or interference scenarios. The model classifies QR and ASCII-coded images ($d_f$ and $a_f$ frames in Table II) with augmented data from rotation and cropping. As shown in Fig. 6, training accuracy starts below 0.5% but improves steadily with more epochs.

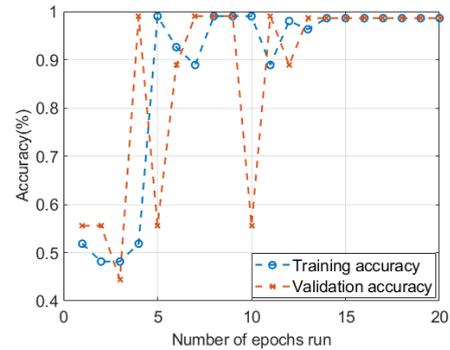

Figure. 6. Training and testing accuracy of data frame detection.

The training curve plateaus after 7 epochs, while the validation curve starts with a low accuracy of 0.444% but quickly reaches 99% within just 5 epochs. The results demonstrate that the model achieves over 98.60% accuracy in detecting and analyzing various transmitted codes, effectively separating interference signals at the receiver.

3) Overhead frame detection

The third experiment evaluates the CNN's ability to distinguish QR-code-based overhead frames from data frames for synchronization. Training on diverse encoded data improves the system's capacity to accurately detect and classify overhead and data frames, ensuring robust synchronization. This enhances S2C performance by strengthening payload classification, QR code detection, and frame synchronization, improving resilience against interference. The model classifies data frames ($d_{f1}$, $d_{f2}$) and overhead frames ($o_f$) as shown in Table II, using augmented data. Fig. 7 shows the training curve, which starts at 0.45% accuracy and reaches 99.60% after 8 epochs. The validation curve follows a similar trend, achieving 99.60% in under 10 epochs.

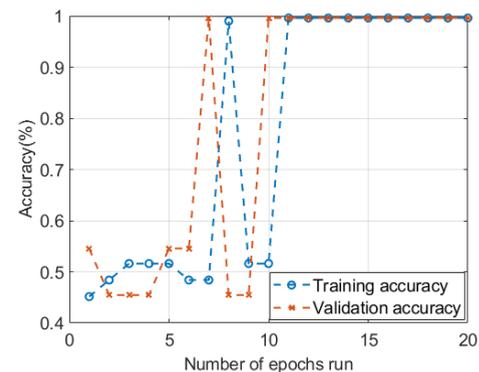

Figure. 7. Training and testing accuracy of overhead detection.

The overall model accuracy for detection and classification of the transmitted codes is represented in Table III.

TABLE III
EXPERIMENTAL RESULTS

| Experimental Investigation | Precision | Recall | F1 - Score | Accuracy (%) |
|---|---|---|---|---|
| Ex1 | 0.980 | 0.986 | 0.985 | 98.60 |
| Ex2 | 0.990 | 0.996 | 0.993 | 99.60 |
| Ex3 | 0.960 | 0.980 | 0.970 | 98.00 |
| Average Model Performance | 0.9767 | 0.987 | 0.982 | 98.74 |

The proposed model can be integrated into the Rx end of S2C for enhanced detection and improved synchronization. This section provides a comparative analysis of the model's performance against existing models, evaluating key metrics such as precision, recall, F1-score, and accuracy, as summarized in Table III. The comparison includes four well-known algorithms: Linear SVM [21], LeNet-5 [22], ImageNet [23], and GoogLeNet [24]. According to the performance data, the proposed CNN model outperforms all other models in overall accuracy and surpasses three models in terms of meantime, as shown in Table IV.

TABLE IV
COMPARISON OF RESULTS WITH OTHER MODELS

| Models | Overall Accuracy (%) | Mean time (ms) |
|---|---|---|
| SVM | 76.32 | 258 |
| LeNet-5 | 78.65 | 210 |
| GoogLeNet | 93.24 | 279 |
| ImageNet | 90.08 | 270 |
| Proposed Method | 98.74 | 237 |

The proposed model demonstrates the highest overall accuracy among all compared models, while achieving the second-best mean time performance when evaluated against SVM, LeNet-5, ImageNet, and GoogLeNet. Despite being a lightweight model, it outperforms the heavier pre-trained models, making it highly efficient and well-suited for real-time S2C-based Android applications. Recent advances in mobile hardware, combined with the efficiency of this lightweight model, have enabled CNNs to run on smartphones with impressive performance and energy efficiency.

### B. SYCHRONIZATION

The S2C system uses asynchronous transmission, where the Tx transmits a frame at time $t_b$. Although there is some propagation delay due to the transmission, it is negligible in S2C camera communication because of the short link distance, light speed, and high camera efficiency. The main source of delay comes from computation at the Rx end. The received signal and its sampled version are expressed as:

$$r(t) = \sum_{k=-\infty}^{\infty} s(k)p(t_b - kF - t_0 - T) + w(t) \quad (1)$$

$$r[n] = r(t)|_{t=nT/Q} \quad (2)$$

Where, $F$ is the frame duration, $s(k)$ is the Tx symbol, $p(t)$ is the filter's impulse response, $w(t)$ is Gaussian noise, $t_0$ is the packet arrival time, and $T$ is the Rx computation time.

Each data frame consists of 4,000 characters, which can be computationally intensive (the maximum QR code capacity is 4,000 characters). The frame rate (fps) for 4,000 characters is around 0.75 fps as represented in Fig.8, especially for devices like the Google Pixel 6 Pro. Due to the smaller payload, overhead frames are easier for the receiver to detect. The CNN model is trained to recognize sync frames, start the stream of frames, and correct any timing offsets.

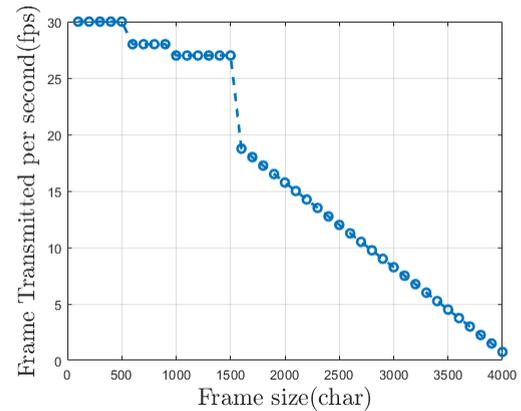

**Figure. 8.** Training and testing accuracy of overhead detection.

The efficiency of the CNN in detecting overhead and data frames is reflected by the ratio x, where, x → 1 for data frames and x<<1 for overhead frames. The system gain with the CNN algorithm is represented by the equation:

$$x(t) = \frac{T - T_{cnn}}{T}, \quad (3)$$

Here, $x(t)$ is the system gain, $T$ is the computation time for a frame, and $T_{cnn}$ is the CNN's processing time. With an overhead frame of 100 characters, the system gain is estimated at 85%, reducing computation time from 33.33 ms to 5 ms by skipping payload processing for sync frames.

While current data rates are limited by smartphone processing speeds, future advancements in mobile hardware are expected to increase the data rate of S2C systems significantly.

### V. CONCLUSION

In this paper, a CNN-based approach was proposed to address the frame identification and synchronization problems in S2C system. The model can deal with raw inputs directly and extract features automatically before performing classification. It has also been proven to be an effective method in real life scenarios. As shown by experimental results, we achieved an overall model accuracy of 98.74% in classification and the system was able to perform with least degradation with small training samples. Moreover, it helps in improving the transmission accuracy, synchronization, practicality, and detection in S2C system. Notably, the comparison results showed that the proposed algorithm outperformed SVM, LeNet, ImageNet and GoogLeNet in classification accuracy. In terms of computation time, the proposed system outperformed all the algorithms but only slightly took more time than state-of-the-art LeNet. The current processing speed of smartphones limits the data rate of the S2C system. As the smartphone technology advances, the data rate of the S2C system is expected to increase. Overall, future advancements in smartphone technology are likely to enhance the performance of the S2C system. The low complexity and less computational capacity of the model have made it preferable to be utilized in S2C system.